\documentclass{article}
\usepackage[preprint]{neurips_2026}
\usepackage{amsmath,amsfonts,amssymb,amsthm}
\usepackage{graphicx}
\usepackage{geometry}
\usepackage{booktabs}
\usepackage{algorithm}
\usepackage{algorithmic}
\usepackage{hyperref}
\usepackage{bm}
\usepackage{cleveref}
\usepackage{subcaption}
\usepackage[table]{xcolor}
\usepackage{tabularx}
\usepackage{tikz}
\usetikzlibrary{positioning,arrows.meta,shapes,calc}
\usepackage{siunitx}
\usepackage{multirow}

\newcommand\figsubref[2]{\hyperref[#1]{\ref*{#1}#2}}

\title{Information Dynamics of Language Communication}

\author{%
  Leonardo S. Goodall\\%
  Calleva Research Centre, University of Oxford, UK\\%
  \And%
  Andrea Luppi\thanks{Shared senior author}\\%
  St John's College, University of Cambridge, UK\\%
  Montréal Neurological Institute, McGill University, Canada\\%
  Centre for Eudaimonia and Human Flourishing, University of Oxford, UK\\%
  \And%
  Pedro Mediano\footnotemark[1]\\%
  Department of Computing, Imperial College London, UK
}

\begin{document}

\maketitle

\begin{abstract}
Quantifying how meaning propagates through communicative exchanges remains underdeveloped in computational linguistics. Here we introduce an information-theoretic framework that quantifies the directed flow of semantic content between interlocutors and decomposes multi-source contributions into redundant, unique, and synergistic components. Our approach leverages large language models as probabilistic estimators of natural language to compute two measures: semantic transfer entropy (STE), which captures directed predictive influence between speakers, and semantic partial information decomposition (SPID), which resolves how multiple sources jointly shape a target's language. Across four experiments we show that the framework detects reduced information flow in cognitively rigid dialogue, captures the dominant role of persuaders in shaping discourse, distinguishes high- from low-quality psychotherapy by the directionality of therapist-client information exchange, and reveals synergistic premise contributions in argumentative essays. This framework opens new avenues for studying information dynamics in digital discourse, pedagogical interactions, clinical dialogues, and any domain in which the structure of linguistic exchange is of research relevance.
\end{abstract}

\section*{Introduction}

Humans use language to communicate. Sharing information that is first held in the privacy of one's mind requires communication, be it via gestural, graphical, linguistic, or other channels. For communication to succeed, information must not only be transmitted but also received, understood, and internalised by the recipient(s), such that the interlocutors come to share a representation of that information \cite{chidichimo_towards_2026}. Yet, capturing and quantifying how meaning propagates through these communicative channels has proved difficult.

Computational linguistics has advanced from surface-level lexical measures \cite{church_word_1989} through distributional semantics \cite{deerwester_indexing_1990} to contextual neural representations \cite{vaswani_attention_2023, devlin_bert_2019}, progressively deepening what can be modelled. Nonetheless, no existing approach provides a principled, unified framework for quantifying how much semantic information flows between interlocutors, in which direction, and how multiple sources jointly contribute to what is said next. Current approaches rely on ad hoc measures, surface lexical features, or correlational summaries that address directionality or multi-source structure separately, but not both in a principled and unified way.

Information theory has already been recognised as a promising and powerful probability-based framework for measuring and modelling social communication and coordination \cite{chidichimo_towards_2026}, offering model-free and data-driven alternatives to assumptive and model-dependent approaches. Shannon himself drew upon natural language as his founding example, modelling English as a stochastic process and demonstrating that the predictability of each successive letter or word can be precisely quantified in bits. Since its inception, the framework has been extended in two directions particularly relevant to the study of communication: transfer entropy \cite{schreiber_measuring_2000}, which quantifies the directed flow of information from one process to another, and partial information decomposition \cite{williams_nonnegative_2010}, which decomposes the information that multiple sources provide about a target into redundant, unique, and synergistic contributions. These measures, however, require well-defined probability distributions over the variables of interest. One natural approach would be to model communicative exchanges as trajectories through the high-dimensional embedding spaces of neural language models \cite{ding_same_2023}, but such representations do not readily yield the discrete probability distributions that information-theoretic quantities demand. Large language models (LLMs) address this directly: trained as next-token predictors over vast text corpora, they provide rich, context-sensitive probability estimates over natural language, lending themselves naturally as the probabilistic backbone for information-theoretic computation \cite{linzen_syntactic_2021, pavlick_semantic_2022}.

Here, we combine these research lines to introduce two novel measures. Semantic transfer entropy (STE) quantifies directed predictive dependence between communicators by comparing a language model's predictions of a target's utterances with and without access to a source's prior content. Semantic partial information decomposition (SPID) extends this to the multi-source case, decomposing how multiple speakers jointly contribute to a target's language into redundant, unique, and synergistic components. We validate both measures across four experiments spanning synthetic conversations with manipulated cognitive styles, a naturalistic persuasion corpus, client–therapist dialogues, and argumentative essays, demonstrating that the framework detects theoretically meaningful variation in each domain. Such a framework is nonetheless portable across any system in which agents exchange meaningful sequential text, with applications spanning clinical dialogue, political and media communication, education, group deliberation, and human-AI interaction.

This work makes the following contributions:
\begin{itemize}
    \item We introduce semantic transfer entropy (STE), an information-theoretic measure of directed semantic influence between communicators implemented via attention masking in LLMs.
    \item We introduce semantic partial information decomposition (SPID), which decomposes the contributions of multiple sources to a target's language into redundant, unique, and synergistic components, and validate it against annotated argumentative structure.
    \item We demonstrate that both measures are sensitive to theoretically meaningful distinctions across domains, including cognitive rigidity, persuasion strategy, therapeutic alliance quality, and premise-claim informational structure.
    \item We show that key findings replicate across three language models spanning two architectures and two scales, supporting the robustness of the framework independent of the choice of estimation model.
\end{itemize}

\section*{Results}

\subsection*{Measuring Semantic Information Flow}

\begin{figure*}[!ht]
    \centering
    \includegraphics[width=1\linewidth]{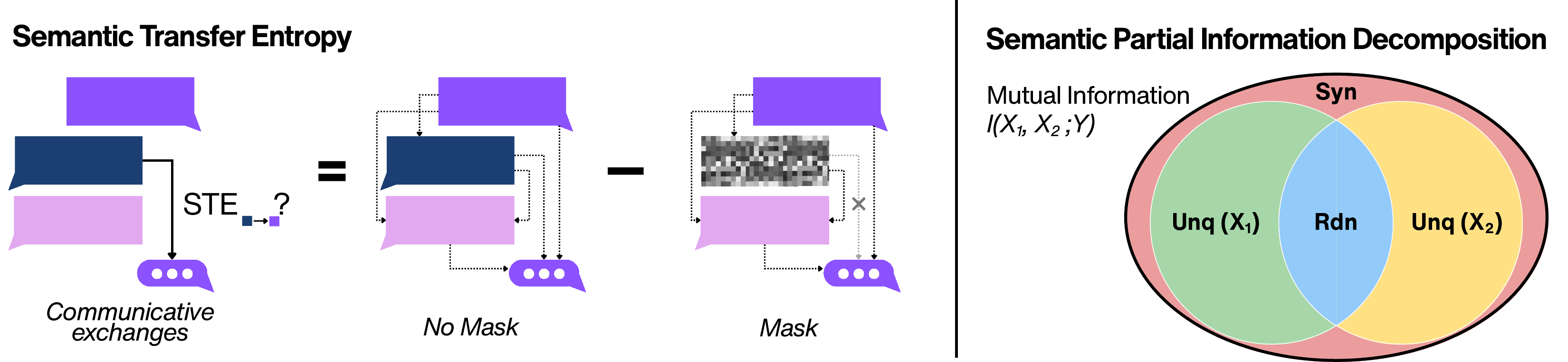}
    \caption{\textbf{Quantifying Semantic Information Flow.} Taking chat logs as an example of communicative exchanges, semantic transfer entropy (STE; left) quantifies the semantic influence a source (dark blue) has on a target (purple's next post). Methodologically, this is implemented by masking attention from the target to the source only. The difference in the target's log-likelihood with and without the mask yields the transfer entropy (in bits). \textit{Note.} Attention from non-targets to the source and from the source to non-target messages are left unchanged, isolating the direct information flow from source to target while preserving the rest of the conversational context. Semantic partial information decomposition (SPID; right) extends this to the multi-source case, decomposing how multiple speakers jointly contribute to a target's language into redundant, unique, and synergistic components.}
    \label{fig:ste_spid}
\end{figure*}

We introduce two complementary measures for quantifying language dynamics. Semantic transfer entropy (STE) captures directed predictive dependence between communicators, building on the information-theoretic measure, transfer entropy \cite{schreiber_measuring_2000}: if person $Y$’s past semantic content improves predictions of person $X$’s future content beyond what $X$’s own history affords, $Y$ exerts directional predictive influence over $X$. LLMs provide a natural estimation vehicle, since they are trained as next-token predictors and afford direct access to conditional text probabilities (Figure \ref{fig:ste_spid}). For any two interacting persons $X, Y$ with prior utterances $X^-, Y^-$, and all remaining conversational context $Z^-$ (i.e. prior utterances from all other participants):
\begin{align}
    STE_{Y\rightarrow X} &= I(Y;X|X^-, Z^-) \label{eq:ste} \\
    &= \mathbb{E}\Big[\log p(X|X^-,Y^-,Z^-)-\log p(X|X^-,Z^-)\Big]. \notag
\end{align}
Here $Y$ is the source whose content is withheld and $X$ is the target whose predictability is assessed. (See Supplementary Information for full derivation and implementation details.)

STE captures pairwise directional dependence, but a person’s next utterance may draw on information uniquely attributable to one partner, shared across multiple partners (redundant), or only accessible when sources are considered jointly (synergistic; \cite{williams_nonnegative_2010, luppi_information_2024}). Semantic partial information decomposition (SPID) resolves these contributions, addressing the question of what source provides what information about the target \cite{griffith_quantifying_2014, griffith_quantifying_2015} (Figure \ref{fig:ste_spid}):
\begin{align*}
    I(Y; X_1,X_2) =\,\, &\text{Rdn}(X_1,X_2 \rightarrow Y) \, + \\
    &\text{Unq}(X_1 \rightarrow Y) + \text{Unq}(X_2 \rightarrow Y) \, +\\
    &\text{Syn}(X_1,X_2 \rightarrow Y)
\end{align*}
We adopt the minimum mutual information (MMI) redundancy functional, which identifies redundancy as the lesser of the two marginal source–target mutual informations \cite{barrett_exploration_2015}; once selected, all remaining atoms follow. Unlike STE (which operates over multi-turn dialogue where intermediate messages respond to the source, necessitating attention masking), SPID applies to configurations where sources are independent texts (e.g., premises supporting a claim), permitting marginalisation by physically omitting a source from the input sequence and yielding true marginal distributions without cascading representational changes (see Supplementary Information for full methodological detail).

For more than two sources, the number of PID atoms grows as 4, 18, and 166 for $n = 2, 3, 4$ \cite{williams_nonnegative_2010}, making a full decomposition unwieldy. The redundancy-synergy index (RSI; also called the whole-minus-sum measure) provides a scalar summary of the redundancy-synergy balance for any $n$ \cite{brenner_synergy_2000, schneidman_synergy_2003, gutknecht_shannon_2025}:
\begin{equation}
    RSI(X_1, \ldots, X_n; Y) = I(Y;X_1,\ldots,X_n) - \sum_{i=1}^{n} I(Y;X_i) \label{eq:rsi}
\end{equation}
Positive RSI indicates synergy dominance (the whole exceeds the sum of parts); negative values indicate redundancy dominance (see Supplementary Information).

\subsection*{Experiment 1: Cognitively rigid conversations are marked by reduced semantic information flow}

Cognitive rigidity—the difficulty in shifting from established patterns of thinking when faced with new information—is associated with reduced perspective-taking and diminished responsiveness to conversational partners \cite{zmigrod_ideological_2025}. As a controlled proof-of-concept, we used synthetic conversations in which cognitive style was experimentally manipulated, enabling a clean test of whether STE is sensitive to theoretically predicted differences in information dynamics. Specifically, we predicted that cognitively flexible speakers would generate stronger directed information transfer than rigid ones. This experiment is intended to demonstrate the sensitivity of STE to manipulated communicative style under controlled conditions, not to validate STE as a measure of cognitive flexibility as a psychological construct.

Source cognitive style was the dominant predictor of STE (Supplementary Table~\ref{tab:source_effect}). A factorial mixed-effects model revealed a highly significant main effect of source type ($F(1, 5995) = 41.07$, $p < .001$): flexible sources generated substantially higher STE ($M = 0.52$, $SD = 0.38$) than rigid sources ($M = 0.46$, $SD = 0.36$). Target type, by contrast, showed no independent effect when source type was controlled ($F(1, 5995) = 0.01$, $p = .920$). The source $\times$ target interaction was significant ($F(1, 5995) = 5.59$, $p = .018$), indicating that the source effect was amplified in flexible-to-flexible pairings.

The overall effect of direction type was highly significant ($F(3, 5995) = 17.11$, $p < .001$; Table~\ref{tab:direction_means}). Flexible-source directions (f$\rightarrow$f and f$\rightarrow$r) showed consistently higher STE than rigid-source directions (r$\rightarrow$f and r$\rightarrow$r), with f$\rightarrow$f showing the highest STE and differing significantly from r$\rightarrow$f ($p < .001$) and r$\rightarrow$r ($p < .001$) after Tukey correction. Critically, f$\rightarrow$r also significantly exceeded r$\rightarrow$f ($p < .001$, $d = 0.18$), confirming that the source's cognitive style drove the effect regardless of the target's style.

Within mixed-dyad conversations, where source and target cognitive style are crossed with speaker position, paired analysis confirmed that flexible sources generated significantly higher STE than rigid sources: flexible-to-rigid STE ($M = 0.51$) exceeded rigid-to-flexible STE ($M = 0.45$) within the same conversations ($t(1999) = -4.05$, $p < .001$, $d = 0.18$). This effect was robust to controlling for speaker position (A vs B; $p = .073$ for position, with the source effect unchanged at $F(1, 3995) = 30.71$, $p < .001$).

\begin{table}[h]
\centering
\caption{Semantic transfer entropy by direction type (source $\rightarrow$ target).}
\label{tab:direction_means}
\begin{tabular}{lcccc}
\hline
Direction & $M$ & $SD$ & $n$ & Contrast vs f$\rightarrow$f \\
\hline
f$\rightarrow$f & 0.538 & 0.386 & 1000 & --- \\
f$\rightarrow$r & 0.513 & 0.377 & 2000 & $p = .301$ \\
r$\rightarrow$f & 0.450 & 0.351 & 2000 & $p < .001$ \\
r$\rightarrow$r & 0.472 & 0.372 & 1000 & $p < .001$ \\
\hline
\end{tabular}
\end{table}

This experiment confirms that STE is sensitive to a manipulated communicative variable: cognitively flexible sources generate substantially higher directed information transfer than rigid ones, regardless of target style. We treat this strictly as a controlled sensitivity demonstration, establishing the measure's discriminative capacity before applying it to naturalistic data in the experiments that follow.

\subsection*{Experiment 2: Persuaders dominate directed semantic information flow}

\begin{figure*}[!ht]
\centering
\includegraphics[width=\linewidth]{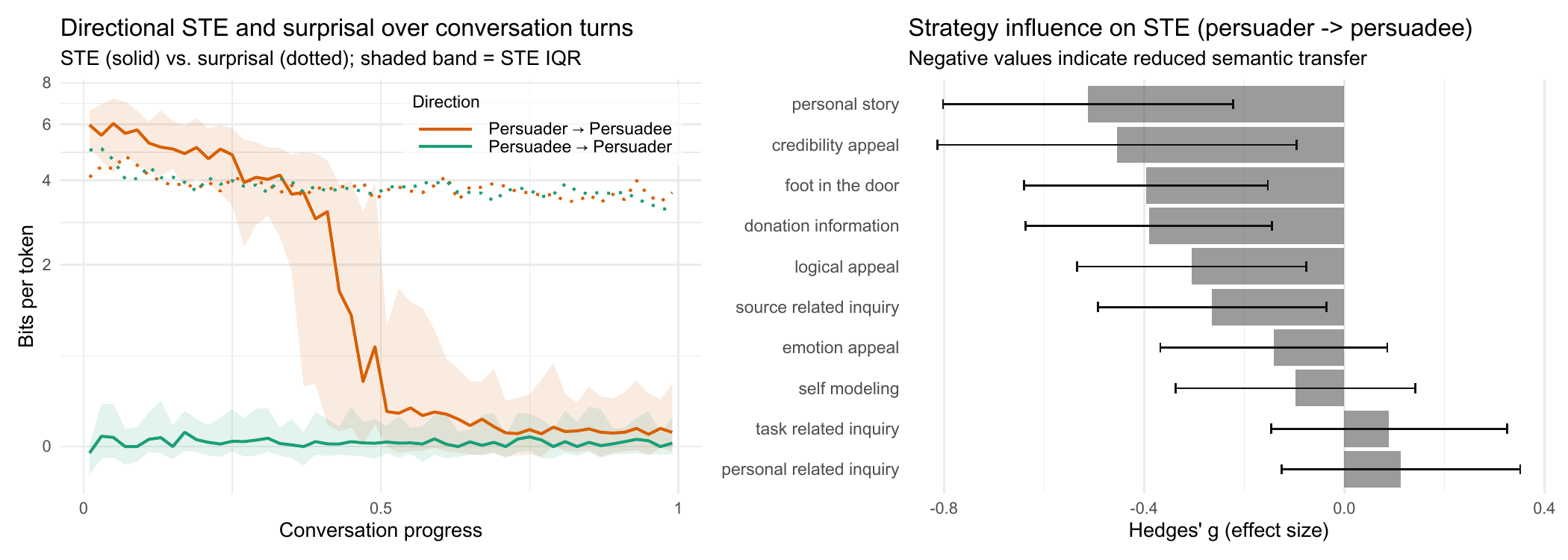}
\caption{STE in persuasion dialogues. Left: Directional STE and surprisal trajectories across normalised conversation progress (solid = STE, dotted = surprisal; shaded band = IQR). Right: Strategy effects on STE (Hedges' $g$); negative values indicate reduced semantic transfer when strategies are employed.}
\label{fig:persuasion_results}
\end{figure*}

Persuasion involves asymmetric information flow: persuaders actively shape discourse while persuadees respond. To test whether STE captures this directional asymmetry and how specific rhetorical strategies modulate information transfer, we analysed the \textsc{PersuasionForGood} corpus \cite{wang_persuasion_2020}, comprising 300 dyadic conversations (10,785 exchanges) in which persuaders solicited charitable donations from persuadees, with each turn annotated for ten persuasion strategies. This experiment demonstrates STE's sensitivity to role structure and strategic communication in naturalistic dialogue.

STE revealed striking directional asymmetry (Fig.~\ref{fig:persuasion_results}). Information transfer from persuader to persuadee (mean = 1.70 bits/token, SD = 1.46) exceeded the reverse direction (mean = 0.05 bits/token, SD = 0.09), with 98\% of conversations showing this pattern. Most notably, the time course, from exposition to mid-conversation, demonstrates a clear descent in STE from persuader to persuadee. This asymmetry remained stable across conversation stages, and was not reducible to simple predictability: the surprisal baseline showed no relationship with donation outcomes and did not attenuate the role effect in mixed-effects models.

Strategy analysis revealed significant negative associations between several persuasion strategies and STE (Fig.~\ref{fig:persuasion_results}, right). Personal stories showed the strongest effect ($g = -0.51$, $p < .001$), followed by credibility appeals ($g = -0.45$, $p < .001$), foot-in-the-door ($g = -0.40$, $p < .001$), and donation information ($g = -0.39$, $p < .001$). Effect sizes were modest ($g \approx -0.3$ to $-0.5$), and the remaining strategies did not reach statistical significance.

These negative effects are consistent with the interpretation that such strategies introduce self-contained content, about the persuader's experiences, credentials, or the cause, that contributes little incremental signal for predicting the persuadee's next turn beyond what is already captured by the persuadee's own conversational history. In other words, this does not imply persuasive ineffectiveness, only reduced predictive coupling about the persuadee's statements. The scope of this interpretation is exploratory, and future work should test it across additional persuasion corpora.

\subsection*{Experiment 3: Higher-quality therapy is characterised by client-driven, not therapist-driven, semantic information flow}
% Experiment 3: Better therapy emerges when clients, not therapists, drive the information flow.
% Experiment 3: Semantic TE reveals asymmetric information flow from therapist to client
% \subsection*{Experiment 3: Better therapeutic alliance is associated with lower therapist-to-client STE}

Motivational interviewing (MI) is a collaborative, client-centred therapeutic style whose defining characteristic is the elicitation of the client's own motivation for change rather than its prescription by the therapist \cite{rollnick_what_1995, miller_motivational_2013}. A large empirical literature has established that therapist verbal behaviour directly shapes therapeutic process: MI-inconsistent behaviours (confrontation, unsolicited advice-giving, and directive steering) are systematically followed by client resistance and sustain talk, whereas MI-consistent behaviours (open questions, reflections, and affirmations) elicit client change talk \cite{moyers_therapist_2006, apodaca_which_2016}. Client change talk, in turn, predicts better behavioural outcomes across addictive behaviours \cite{moyers_-session_2009, magill_meta-analysis_2018}, establishing a causal chain from therapist verbal style through client speech to outcome (e.g., \cite{miller_enhancing_1993}). We therefore hypothesised that therapist-to-client STE would be lower in high-quality sessions, where the therapist follows the client's lead rather than directing it, than in low-quality sessions characterised by therapist-dominated steering. To test this, we analysed the AnnoMI corpus \cite{wu_anno-mi_2022}, comprising 133 naturally occurring MI sessions (110 high-quality, 23 low-quality) annotated by experienced MI practitioners.

The results confirmed the hypothesis. Therapist-to-client STE was substantially higher in low-quality sessions ($M = 1.62$, $SD = 1.66$) than in high-quality sessions ($M = 0.76$, $SD = 0.90$; Welch's $t(24.8) = 2.40$, $p = .024$, $d = -0.80$). The directional asymmetry (STE$_{\text{therapist} \to \text{client}}$ $-$ STE$_{\text{client} \to \text{therapist}}$) was similarly elevated in low-quality sessions ($M = 1.43$ vs.\ $M = 0.46$; $t(27.0) = 2.55$, $p = .017$, $d = -0.72$). A weak but significant negative correlation further linked therapist-to-client STE with the proportion of client change talk across sessions ($r = -.18$, $p = .044$), suggesting that sessions in which therapists generated more directed information flow were also sessions in which clients produced less autonomous change talk.

These findings reveal that STE captures a theoretically meaningful dimension of therapeutic discourse. In low-quality MI, the therapist dominates the informational structure of the exchange (confronting, advising, and directing), rendering the client's subsequent utterances highly predictable from the therapist's prior content. High-quality MI therapists, by contrast, may produce lower STE precisely because they follow the client's lead potentially as their reflections and open questions are shaped by what the client has said, making the conversational influence bidirectional and informationally symmetric. Finally, with respect to the change talk correlation, where therapist informational dominance and client autonomous motivation generation are in tension, STE indexes this tension without access to any coded behavioural categories.

\subsection*{Experiment 4: Argumentative premises contribute synergistic information about claims}

Natural language often emerges from multiple sources. For example, arguments build claims from several premises, and speakers respond to multiple prior utterances. When two sources jointly predict a target, their contributions may be redundant (overlapping), unique (exclusive to one source), or synergistic (the whole exceeds the sum of the parts). We applied SPID to the Argument Annotated Essays corpus (AAE-v2; \cite{stab_parsing_2016}), comprising 402 persuasive student essays with annotated claims, premises, and support relations, focusing on the 325 triplets in which claims were supported by exactly two premises.

\begin{table*}[h]
\centering
\small
\caption{Example SPID decompositions from the AAE corpus. Each example shows a claim and its two supporting premises, with the resulting information atoms (bits per token). Red.\ = Redundancy; Unq.\ = Unique; Syn.\ = Synergy.}
\label{tab:aae_examples}
\begin{tabular}{p{5.8cm}p{4.5cm}ccc}
\toprule
Claim & Premises & Unq. & Red. & Syn. \\
\midrule

\multirow{2}{5.8cm}{\emph{``keeping our pace higher is important''}}
& \scriptsize P1: avoiding sluggishness increases success probability
& $-$0.78 & \multirow{2}{*}{0.00} & \multirow{2}{*}{\textbf{2.61}} \\
& \scriptsize P2: high pace prevents laziness
& 0.30 &  &  \\
\midrule

\multirow{2}{5.8cm}{\emph{``it can do good to protect the endangered animals''}}
& \scriptsize P1: advanced methods to care for animals
& \textbf{2.08} & \multirow{2}{*}{0.00} & \multirow{2}{*}{0.00} \\
& \scriptsize P2: agriculture contributes to economy
& $-$0.20 &  &  \\
\midrule

\multirow{2}{5.8cm}{\emph{``dancing is important in every culture''}}
& \scriptsize P1: dancing shows cultural civilization
& 0.00 & \multirow{2}{*}{\textbf{3.76}} & \multirow{2}{*}{0.88} \\
& \scriptsize P2: people use dancing to entertain themselves
& 0.10 &  &  \\
\bottomrule
\end{tabular}
\end{table*}

Table \ref{tab:aae_examples} presents representative examples illustrating the range of SPID decompositions observed. The first example shows the highest synergy: two premises addressing distinct aspects of pace (success probability and laziness prevention) that jointly predict a claim about the importance of maintaining high pace. The second example demonstrates maximal unique information: one premise directly discusses animal protection methods while the other (about agriculture and economy) contributes nothing relevant. The third example shows the highest redundancy, where both premises convey culturally-relevant aspects of dancing that overlap in predicting the claim.

\begin{figure*}[!ht]
    \centering
    \includegraphics[width=1\linewidth]{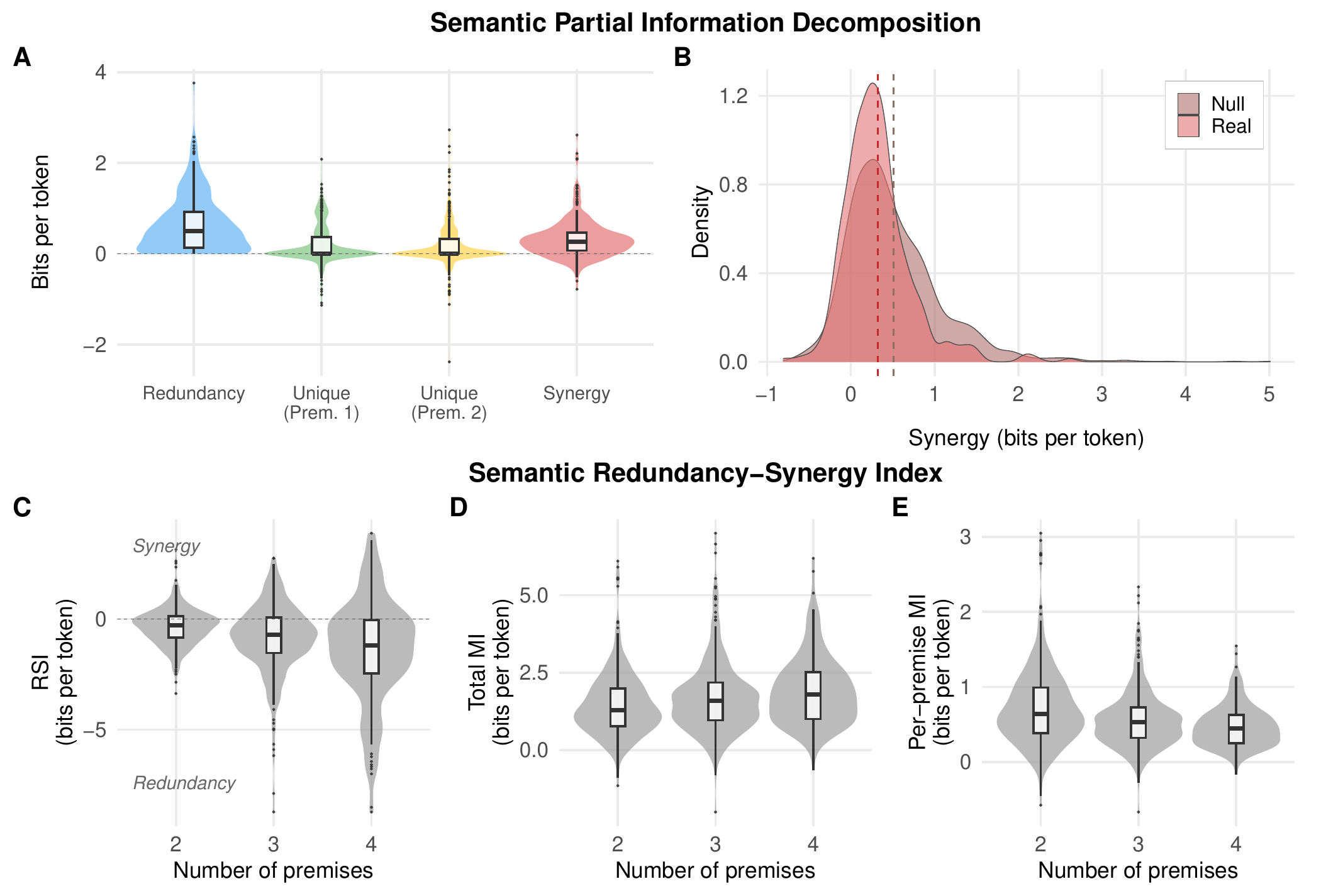}
    \caption{\textbf{Two-premise arguments exhibit positive synergy, while aggregating more premises produces growing redundancy.} (A) Distribution of SPID atoms across 325 two-premise claim triplets. Synergy is positive on average, indicating super-additive predictive information. (B) Synergy distributions for real and shuffled premise-claim pairings ($p < .001$). (C) RSI becomes increasingly negative as premise count grows from 2 to 4 ($p < .001$), indicating that the redundancy-synergy balance shifts toward redundancy dominance with more premises. (D) Total mutual information (joint MI) increases with premise count but grows sublinearly. (E) Per-premise information yield (joint MI / $n$) decreases with more premises, indicating diminishing returns ($p < .001$).}
    \label{fig:aae_pid}
\end{figure*}

Synergy was positive on average (M = 0.32, SD = 0.42 bits/token; Figure \figsubref{fig:aae_pid}{A}), significantly above zero ($t(324) = 13.90$, $p < .001$, $d = 0.77$), indicating that premises jointly provided more predictive information about claims than the sum of their individual contributions. Redundancy was moderate (M = 0.63, SD = 0.62), reflecting shared predictive content between premises. Unique information was modest and symmetric: Premise 1 contributed M = 0.20 (SD = 0.43) and Premise 2 contributed M = 0.19 (SD = 0.48). Real premise-claim pairings showed significantly higher redundancy than shuffled pairings (real: M = 0.63; null: M = 0.12; $t(1948) = 14.62$, $p < .001$, $d = 1.42$; Figure \figsubref{fig:aae_pid}{B}), confirming that premises supporting the same claim share topically relevant information, whereas random premise pairs lack this coherence. Correlations with semantic similarity (cosine similarity of sentence embeddings; \cite{gao_simcse_2021, reimers_sentence-bert_2019}) further corroborated the decomposition: redundancy was positively correlated with premise similarity ($r = .23$, $p < .001$), synergy showed a marginal negative correlation ($r = -.10$, $p = .06$; \cite{caprioglio_synergistic_2026}), and unique information was strongly correlated with premise-claim similarity (Premise 1: $r = .40$; Premise 2: $r = .41$; both $p < .001$). These patterns confirm that SPID captures meaningful semantic relationships rather than superficial text statistics (see Supplementary Information for the full null model analysis and CCS/MMI comparison).

To examine how informational structure changes as more premises support a claim, we extended the analysis using RSI (Equation~\ref{eq:rsi}). We extracted all claims supported by exactly 3 or 4 premises from the AAE corpus (307 and 159 samples, respectively) and computed RSI alongside the existing $n = 2$ samples ($N = 791$ total). RSI was negative at every premise count and grew more negative with $n$: $M = -0.34$ for $n = 2$, $M = -0.87$ for $n = 3$, and $M = -1.41$ for $n = 4$ (Kruskal--Wallis $H(2) = 43.54$, $p < .001$; all pairwise comparisons significant after Holm correction), indicating that the redundancy-synergy balance shifted toward redundancy dominance as more premises supported a claim. The proportion of synergy-dominated samples (RSI $> 0$) declined from 32\% ($n = 2$) to 27\% ($n = 3$) to 24\% ($n = 4$; Cochran--Armitage trend: $\chi^2 = 4.18$, $p = .04$). Per-premise information yield (joint mutual information divided by the number of premises) also decreased across groups ($M = 0.73$, $0.57$, $0.48$ for $n = 2, 3, 4$; linear slope $= -0.13$, $p < .001$), indicating diminishing returns: each additional premise contributes progressively less novel predictive information about the claim. Total mutual information grew sublinearly with premise count (Figure \figsubref{fig:aae_pid}{C--E}).

SPID decomposed the information structure of argumentative essays and showed that two-premise arguments exhibit positive synergy, with premises jointly providing more predictive information about the claim than the sum of their individual contributions. Extending to 3- and 4-premise arguments with RSI, the redundancy-synergy balance shifted progressively toward redundancy as more premises were added, alongside diminishing per-premise information yield. Aggregating more premises about a fixed claim thus produces increasingly overlapping rather than complementary evidence.

All key effects across Experiments 1--4 replicated in direction and significance across three models spanning two architectures and two scales (LLaMA~3.2-3B, Phi-3-mini-4k, Mistral-7B-v0.3; Supplementary Tables~S2--S5).

\section*{Discussion}

The present work introduced two information-theoretic tools for studying language communication dynamics at the semantic level: STE for quantifying directed information flow between interlocutors, and SPID for decomposing multi-source contributions into redundant, unique, and synergistic components. Four experiments, each with a distinct evidential role, support the framework. Experiment 1 serves as a controlled proof-of-concept: by manipulating cognitive style in synthetically generated conversations, it demonstrates that STE is sensitive to a theoretically meaningful variable under conditions of experimental control, establishing the method's discriminative capacity before applying it to more complex settings. Experiment 2 provides empirical evidence in naturalistic human dialogue, capturing asymmetric directional influence in a real persuasion corpus. Experiment 3 demonstrates that STE indexes a clinically meaningful dimension of therapeutic quality in motivational interviewing, without access to any coded behavioural categories. Experiment 4 offers corpus-based validation of SPID against annotated argumentative structure. Together, they demonstrate that these measures operate directly on the full semantic content of utterances rather than on superficial extracted features or predefined lexical categories.

\subsection*{Related work}

The field of computational linguistics has made significant strides in modelling the dynamics of human communication at the lexical, syntactic, and semantic levels of representation. Early computational approaches focused on surface-level linguistic features such as simple word frequencies shared between conversants \cite{nenkova_high_2008}, or the degree to which speakers mirror function-word choices (pronouns, articles, prepositions, conjunctions, and auxiliary verbs) \cite{danescu-niculescu-mizil_chameleons_2011}, to simply correlate with conversation-level metrics such as perceived naturalness of conversation, task success, and even power asymmetries (e.g., \cite{danescu-niculescu-mizil_chameleons_2011, chang_convokit_2020}). Dictionary-based tools such as Linguistic Inquiry and Word Count \cite{tausczik_psychological_2010} extended this paradigm by mapping words onto psychologically meaningful categories, while distributional methods such as Latent Semantic Analysis \cite{landauer_solution_1997} and topic models \cite{blei_latent_2003} moved beyond raw counts to capture latent co-occurrence structure. However, these approaches remain constrained to predefined lexical categories or static distributional statistics and thus sensitive to context and word order, for example, limiting their ability to capture context-dependent meaning. The emergence of neural word embeddings \cite{mikolov_efficient_2013}, contextual representations \cite{devlin_bert_2019}, and large-scale autoregressive language models has since opened the possibility of modelling language to a depth that earlier feature-engineering approaches could not reach.

Trained on large text corpora in a self-supervised manner, LLMs learn the statistical regularities of natural language and embed them into high-dimensional semantic spaces \cite{linzen_syntactic_2021, pavlick_semantic_2022}, such that the word `charge', for example, has distinct representations depending on whether it appears in the context of electricity, economic transactions, or criminal law. Indeed, these models jointly encode morphological, syntactic, semantic, and pragmatic dimensions into a unified representational space. In this way, language can be represented mathematically, allowing communicative exchanges to be modelled as high-dimensional trajectories that reveal fine-grained variations in meaning, context, and conceptual relationships, features previously inaccessible to quantitative analysis. Indeed, some have remarked on the similarity in human language processing and LLM computational principles \cite{kumar_shared_2024, goldstein_shared_2022, heilbron_hierarchy_2022, zada_shared_2024}.

Researchers have begun to exploit these representations to study language dynamics beyond static text classification. Soni and colleagues \cite{soni_predicting_2022} used contextual embeddings to detect semantic shifts in scientific discourse (e.g., the word `attention' migrating from its generic sense to its deep-learning definition) and showed that such short-term linguistic influence predicts long-term citation impact. At the level of public discourse, Ding and colleagues \cite{ding_same_2023} derived a measure of semantic polarity from contextual embeddings and applied Granger causality to establish that divergence in how broadcast news outlets frame the same keywords forecasts subsequent polarisation on social media. Others have operationalised interpersonal linguistic coordination as geometric distance in embedding space, showing that the proximity of successive turns predicts therapist empathy and therapeutic outcomes \cite{nasir_modeling_2019}. 

Most strikingly, Zada and colleagues \cite{zada_shared_2024} used GPT-2 embeddings to trace meaning transfer in natural conversation at the neural level, demonstrating that the same contextual representations that emerge in the speaker's brain before articulation re-emerge, word by word, in the listener's brain—providing direct neural evidence that language models capture the representational space through which meaning is transmitted between minds. Together, these studies demonstrate the power of LLM-derived representations for characterising language dynamics at the semantic level, yet each relies on ad hoc measures of influence (embedding distances, Granger causality, or correlation), none of which afford a principled, unified account of semantic information directionality and multi-source contributions. Here, STE and SPID address this limitation.

An early attempt to apply information-theoretic measures to linguistic influence can be attributed to Ver Steeg and Galstyan \cite{steeg_information-theoretic_2013}, who introduced ``content transfer'' (an adaptation of transfer entropy) to measure influence between Twitter users based on the co-movement of topic distributions derived from Latent Dirichlet Allocation. However, applying TE directly in this way is far from trivial and methodologically contentious as semantic trajectories in embedding spaces are high-dimensional (and thus sparse), temporally interdependent, and noisy, raising challenges of stability, efficiency, and scalability. More recently, Kiddle and colleagues \cite{kiddle_network_2024} extended TE to study the propagation of discursive toxicity on social media, using TE to construct ``toxicity networks'' that reveal how individual users influence one another's toxic output over time. While methodologically sophisticated, this approach again operates on extracted features (toxicity scores based on text annotation) rather than the full semantic content of utterances. STE instead works directly on the semantic content itself, leveraging both the context-rich modelling of LLMs and their internal probability-based computations.

Applications of PID have ranged from cellular automata and artificial neural networks to gene regulatory interactions, socioeconomic data, and neuroscience \cite{finn_quantifying_2018, rosas_information-theoretic_2018, beer_information_2015, tax_partial_2017, makkeh_general_2025, varley_untangling_2022, cang_inferring_2020}. The temporal extension of PID, integrated information decomposition, has further enabled the decomposition of information storage, transfer, and integration in complex dynamical systems, revealing previously unreported modes of collective information flow across over 1,000 biological, physical, social, and synthetic systems \cite{mediano_toward_2025}. Indeed, this measure has been applied to LLMs, uncovering a ``synergistic core'' in middle transformer layers that mirrors the informational architecture of the human brain \cite{urbina-rodriguez_brain-like_2026}. Here, we have presented how information decomposition extends to the semantic case via SPID.

\subsection*{Envisioned applications}

The generality of this framework is one of its principal contributions. Where existing approaches to language dynamics rely on ad hoc measures, surface lexical features, or correlational summaries, the properties of the information-theoretic framework presented here make the framework portable across any setting where meaningful sequential text is exchanged.

In clinical settings, STE could track the bidirectional flow of semantic influence between therapist and patient across sessions, potentially serving as an objective marker of therapeutic alliance or disengagement, complementing recent efforts to infer alliance computationally from session transcripts \cite{chen_characterizing_2018, na_survey_2025}. In political and media communication, the directed and asymmetric nature of STE makes it well-suited for tracing how semantic framings propagate from institutional sources (e.g., news outlets, political leaders) to public discourse, extending recent correlational work on semantic polarisation \cite{ding_same_2023} with a principled directional measure (e.g., \cite{kiddle_network_2024}). In education, SPID could decompose how multiple instructional sources (textbooks, lectures, peer discussion) contribute redundant, unique, or synergistic information to student understanding. More broadly, any communicative system in which multiple agents produce sequential text is amenable to analysis with these tools (e.g., group deliberation \cite{chirigati_collective_2024}, online forums, human-AI collaboration), offering a unified information-theoretic lens where previously only ad hoc measures were available.

The information-theoretic framework also invites connections beyond interpersonal communication \cite{chidichimo_towards_2026}. A growing body of work in mechanistic interpretability traces how semantic representations propagate through the internal layers of transformer models \cite{ali_entropy-lens_2026}, using techniques such as activation patching and causal tracing to identify which components contribute to specific outputs \cite{bereska_mechanistic_2024, heimersheim_how_2024}. Indeed, partial information decomposition has already been applied to probe the informational architecture of neural networks \cite{tax_partial_2017} and, in neuroscience, to characterise how brain regions integrate and broadcast information \cite{luppi_information_2024}. Future work could explore whether the same decompositions (redundancy, unique, synergy) can characterise how multiple attention heads or layers jointly contribute to a model's predictions, bridging the study of human communication dynamics and AI interpretability.

\subsection*{Structural and semantic information}

A foundational clarification concerns the relationship between our measures and the concept of semantic information. Dretske \cite{dretske_knowledge_1986} distinguished structural information---Shannon's quantitative framework of uncertainty reduction---from semantic information: what a signal specifies about a particular state of affairs. Our framework operates, strictly speaking, on the structural side of this divide: STE and SPID quantify how much uncertainty one speaker's utterances resolve about another's, measured in bits via conditional log-likelihoods. The ``semantic'' in semantic transfer entropy refers not to a formal theory of meaning but to the fact that the underlying probability estimates are furnished by a language model trained on the statistical regularities of meaningful language, and thus implicitly conditioned on morphological, syntactic, and pragmatic structure in a way that Shannon's original character-level analyses were not. Although STE and SPID are formally structural quantities, the LLM's probability landscape ensures that the predictive coupling we measure operates over contextualised semantic representations rather than raw symbol frequencies.

\subsection*{Limitations and Future Work}

Several limitations warrant discussion. First, all information-theoretic estimates are conditioned on the particular language model used, and absolute values are not comparable across models with different tokenisation or calibration (\cite{guo_calibration_2017, oh_why_2023}; temperature scaling can substantially alter surprisal estimates \cite{liu_temperature-scaling_2024}, and miscalibration does not resolve with scale \cite{cao_entropy_2026}). However, given that all key effects across Experiments 1--4 replicated across the three models, it may be fair to conclude that the findings reflect properties of the data rather than the estimation instrument. We thus recommend holding the model constant across conditions within a study.

Second, STE uses attention masking rather than physical omission because omitting a conversational participant would alter positional encodings and hidden states of all downstream messages, conflating direct source contribution with cascading representational change. Masking instead preserves the full sequence and severs only the relevant attention edges, making it the structurally appropriate choice for multi-turn dialogue. The complementary concern that masked-but-present tokens might introduce indirect leakage does not apply in the same way to SPID, which uses physical omission precisely because its sources are independent texts. We empirically compared both methods on Experiment 4 and show in the Supplementary Information that masking and omission are not interchangeable.

Third, because we compute pointwise (per-token) information quantities, individual observations can be negative even when the expected value is positive, as seen in the synergy estimates for Experiment 4. Aggregation over sufficient data mitigates this, but researchers applying these tools to small samples should interpret pointwise values with caution. 

Fourth, the computational cost of running LLM forward passes for each conditioning set scales with the number of sources and messages, which may limit applicability to very large corpora or real-time analysis without further engineering optimisation. 

Finally, while STE measures directed predictive dependence, any causal interpretation rests on the same assumptions as Granger causality \cite{shojaie_granger_2022}: (i) \emph{stationarity}—the joint process must be stationary across the conversation, an assumption that natural dialogues, which evolve in topic and register, may violate; (ii) \emph{no hidden confounders}—unobserved variables such as shared context, topic, or speaker background must not simultaneously drive both parties' language; and (iii) \emph{correct model specification}—the LLM's predictive distribution must adequately capture the relevant statistical dependencies. To the extent that these conditions hold, higher STE supports an inference of directional influence; where they are uncertain, STE is better interpreted as a measure of directed predictive coupling. We retain causal language in the discussion of applications where these assumptions are plausible, while using the term ``directional'' or ``predictive'' in more agnostic contexts throughout the paper.

A further constraint applies to Experiment 1, specifically; that is, the conversations themselves were generated by an LLM rather than produced by human speakers with independently verified cognitive profiles, so the result establishes the discriminative capacity of STE rather than its psychological validity. Indeed, LLM-generated text may exhibit surface regularities (e.g., lexical diversity or contextual adaptability) that correlate with the generation condition label independently of whether they reflect genuine cognitive flexibility. Validating these patterns in naturalistic human conversations with independent assessment of cognitive flexibility remains for future work.

\subsection*{Final Remarks}

We presented two novel information-theoretic measures: semantic transfer entropy and semantic partial information decomposition. These measures leverage LLMs to quantify how meaning flows between communicators and how multiple sources jointly shape a target's language. Across cognitive rigidity, persuasion, and argumentation, these tools detected directed influence asymmetries, strategy-dependent information dynamics, and the redundant, unique, and synergistic structure of multi-source communication, phenomena that surface-level and correlational approaches cannot access. By uniting Shannon's mathematical theory of communication and re-attending to his first linguistic exposition—this time with the semantic representational capacity of modern language models—this work offers a general-purpose, principled framework for studying any system in which agents exchange meaningful sequential text: be it from clinical and political communication to education and human-AI interaction.

\section*{Methods}

\subsection*{Experiment 1}

We generated a synthetic corpus of 3,000 dyadic conversations using GPT-5-nano (snapshot \texttt{gpt-5-nano-2025-08-07}; \cite{openai_gpt-5_2025}). Each conversation comprised 20 turns between two agents (A and B) who began with opposing positions on a given topic. Conversations were generated across four experimental conditions: \emph{rigid-rigid}, \emph{flexible-flexible}, \emph{rigid-flexible} (A rigid, B flexible), and \emph{flexible-rigid} (A flexible, B rigid). The two mixed conditions orthogonalise cognitive style and speaker position (A/B), enabling clean separation of source and target effects on STE. Within each condition, conversations spanned five topics (art, politics, food, science, morality), with 100 conversations per topic for homogeneous conditions and 200 per topic for mixed conditions. The generation prompt defined cognitive rigidity as ``the difficulty in shifting from established patterns of thinking, feeling, or behaving when faced with new information or changing circumstances,'' and cognitive flexibility as ``the ability to adapt thinking and behavior in response to new information or environmental demands.'' Agents were instructed to produce 2--3 sentences per turn. In both mixed conditions, the starting speaker alternated across iterations to control for turn-order effects.

We computed STE bidirectionally (A$\rightarrow$B and B$\rightarrow$A) for each conversation using LLaMA 3.2-3B \cite{grattafiori_llama_2024} with 4-bit quantisation.\footnote{A potential concern is circularity when using a language-model-based method to analyse language-model-generated data. In our case, however, the generative model (GPT-5-nano) and the model underlying our information-decomposition pipeline (Llama 3.2-3B) belong to different model families with different architectures and training data, so the analysis cannot, by and large, recapitulate idiosyncratic artefacts of the generator.} A lag window of 5 turns was used, and we excluded turns 1--5 from analysis to ensure stable estimates (yielding 15 analysable turns per conversation). For each conversation, we obtained turn-level STE values and then averaged across turns 6--20 to obtain conversation-level metrics.

We classified each directional STE measure by the cognitive style of source and target: rigid-to-rigid (r$\rightarrow$r), rigid-to-flexible (r$\rightarrow$f), flexible-to-rigid (f$\rightarrow$r), and flexible-to-flexible (f$\rightarrow$f). This yielded 6,000 conversation-direction observations (1,000 each for r$\rightarrow$r and f$\rightarrow$f from homogeneous dyads; 2,000 each for r$\rightarrow$f and f$\rightarrow$r from mixed dyads, with each direction drawn equally from both speaker positions).

Statistical analysis employed linear mixed-effects models with conversation and topic as crossed random effects. We tested (i) the main effects of source and target cognitive style on STE via a source $\times$ target factorial model, (ii) the overall effect of direction type, and (iii) within-dyad paired comparisons in the mixed conditions where source and target effects are orthogonal. Post-hoc pairwise comparisons used Tukey-corrected $p$-values. Effect sizes are reported as Cohen's $d$.

\subsection*{Experiment 2}

We analysed the \textsc{PersuasionForGood} corpus \cite{wang_persuasion_2020}, comprising 300 dyadic conversations (10,785 exchanges) where persuaders solicited charitable donations from persuadees. In the corpus, each turn was annotated with ten persuasion strategies (see Table \ref{tab:persuasion_strategies}). Donation outcomes were recorded but proved unsuitable for correlation analysis due to heavy zero-inflation (61\% unchanged donations).

We computed STE bidirectionally for each conversation using LLaMA 3.2-3B with 4-bit quantisation and a lag window of 16 turns (selected via surrogate-corrected optimisation; see Supplementary Information). For each target utterance, STE was computed as the difference in predictive log-likelihood with versus without access to the source user's recent posts, normalised to bits per token.

Statistical analysis employed linear mixed-effects models with conversation-level random intercepts, controlling for turn length (quadratic effect of inverse token count) and surprisal (to benchmark against simple predictability). Strategy effects were quantified using Hedges' $g$. To assess behavioural relevance, we fitted logistic regression predicting donation ($>$\$0) from STE, strategy counts, and psychological covariates identified in \cite{wang_persuasion_2020}.

\subsection*{Experiment 3}
We analysed the AnnoMI corpus \cite{wu_anno-mi_2022}, a collection of 133 naturally occurring MI sessions (110 high-quality, 23 low-quality) annotated by experienced MI practitioners. STE was computed bidirectionally for each session using a lag window of 5 turns.  We compared therapist-to-client STE and overall directional asymmetry between quality groups using Welch's $t$-tests, and correlated therapist-to-client STE with the proportion of client change talk across sessions using Pearson's $r$.

\subsection*{Experiment 4}

We analysed the Argument Annotated Essays corpus (AAE-v2; \cite{stab_parsing_2016}), comprising 402 persuasive student essays sourced from essayforum.com with annotated argument components (major claims, claims, premises) and argumentative relations (support or attack). We focused on claims supported by exactly two premises, yielding 325 triplets from 228 essays. This structure mirrors the theoretical setup of SPID: two source texts (premises) jointly and individually contributing predictive information about a target text (claim).

For each triplet, we designated the two supporting premises as sources $X_1$ and $X_2$ and the claim as target $Y$. We computed SPID atoms (redundancy, unique$_1$, unique$_2$, synergy) using the MMI redundancy functional, with all information quantities derived from LLaMA 3.2-3B log-likelihood differences under physical omission of the excluded source (see \emph{Semantic Information Decomposition} above). Text lengths averaged 88 characters for claims (SD = 37), 114 for Premise 1 (SD = 52), and 106 for Premise 2 (SD = 47).

As a unit test, we first validated that the SPID implementation correctly recovers known information-theoretic ground truths. We trained small 2-layer transformers on canonical two-input logic gates (AND, OR, XOR, COPY, UNIQ) and confirmed that SPID recovers the expected decompositions: XOR yields maximal synergy ($>0.8$ bits) and near-zero redundancy; COPY yields maximal redundancy and near-zero synergy; and UNIQ concentrates information in one source. All gates achieved $>95\%$ accuracy before PID computation. While this test uses binary synthetic data rather than natural language, it confirms that the implementation correctly identifies the theoretical ground truths that natural language distributions can only approximate.

To validate that SPID estimates reflect genuine argumentative structure rather than generic text statistics, we constructed a permutation null model: for each real sample, we paired the claim with two premises drawn from \emph{other} essays, breaking the argumentative relationship while preserving marginal distributions. Five permutations per sample yielded 1,625 null triplets. We also correlated SPID atoms with semantic similarity between premises (cosine similarity of sentence embeddings from all-MiniLM-L6-v2) as an external validity check.

Finally, to extend the analysis beyond two sources, we additionally extracted claims supported by exactly 3 or 4 premises (307 and 159 samples, respectively) and computed RSI (Equation~\ref{eq:rsi}) across all three premise counts ($N = 791$).

\section*{Code Availability}

The framework introduced in this paper is available as \textsc{\textbf{PSIDyn}: \textbf{P}ython \textbf{S}emantic \textbf{I}nformation \textbf{Dyn}amics}, an open-source Python package at \url{https://github.com/LeoGoodall/PSIDyn}.

\onecolumn

\newpage 
\newpage

\section*{Supplementary Information}

\subsection*{Semantic Transfer Entropy}

Large language models (LLMs) offer a natural way to estimate STE, since they are trained explicitly as next-token predictors over vast corpora and thus provide direct access to conditional probabilities of text. By treating LLMs as probabilistic sequence models, we can compute STE in terms of conditional log-likelihoods under different conditioning sets, thereby quantifying how much a source user's posts improve prediction of a target user's subsequent text without requiring an intermediate semantic representation or restrictive parametric assumptions. STE quantifies the directed predictive influence of source $Y$ on target $X$ by comparing predictions under a counterfactual conditioning that withholds information from $Y$: the subscript $Y\rightarrow X$ denotes the direction of influence, where $Y$ is the source whose content is withheld and $X$ is the target whose predictability is assessed, consistent with the conditioning on $X$'s own history $X^-$.

We quantify directed linguistic influence, or \textit{semantic transfer entropy} (STE), from a source $Y$ to a target user $X$ in a time-ordered thread by estimating the STE at horizon $\Delta=1$, i.e. how much $Y$’s prior posts improve predictions of $X$’s very next post. Let $X_t$ denote the tokens of $X$'s next post at time $t$, $X_{t-k:t-1}$ the tokens in $X$'s $k$ most recent prior posts (simplified to $X_{t-}$), $Y_{t-}$ the tokens in $Y$'s $\ell$ prior posts within a lag window, and $Z_{t-}$ all other users' tokens. The population quantity of interest is the partial STE
\begin{equation*}
STE_{Y\to X}(\tau) = \, \mathbb{E}\Big[ \log p\big(X_t \mid X_{t-}, Y_{t-}^{(\tau)}, Z_{t-}\big) - \log p\big(X_t \mid X_{t-}, Z_{t-}\big) \Big].
\end{equation*} where $Y_{t-}^{(\tau)}$ restricts $Y$'s past to a lookback window of duration $\tau$. Intuitively, $T_{Y\to X}$ measures how much information the recent content of $Y$ provides about $X$'s next post, beyond $X$'s own history and all other observed context $Z$.

We operationalise these conditionals with a pretrained autoregressive transformer $p_\theta$ and realise the conditioning difference by modifying the attention graph available to $X_t$'s tokens. Let the concatenated thread up to (but not including) $X_t$ be tokenised into a sequence $c_{1:n}$ with a standard causal attention mask $M\in\{0,1\}^{n\times n}$, where $M_{ij}=1$ \emph{iff} token $i$ may attend to token $j\le i$. Let $\mathcal{I}_X$ be the index set of tokens belonging to $X_t$, and let $\mathcal{J}^{(\tau)}_Y$ be indices of tokens authored by $Y$ within the lag window $[t-\tau,t)$. We form a “without-$Y$” mask $\tilde M^{(\tau)}$ that blocks all edges from $X$'s future tokens to $Y$'s lagged tokens while leaving all other edges unchanged:
\[
\tilde M^{(\tau)}_{ij} \;=\;
\begin{cases}
0, & i\in \mathcal{I}_X,\; j\in \mathcal{J}^{(\tau)}_Y,\\
M_{ij}, & \text{otherwise}.
\end{cases}
\]
This preserves the exact token sequence, positions, and formatting; only the accessible information set changes, aligning the estimator with the definition of conditional probabilities.

Given a fixed input sequence $c_{1:n}$ and an attention mask $A\in\{M,\tilde M^{(\tau)}\}$, the model-induced log-likelihood of $X$'s next post factors tokenwise as
\[
\log p_\theta\!\left(X_t \mid c_{1:n}; A\right) \;=\; \sum_{i\in \mathcal{I}_X} \log p_\theta\!\left(x_i \mid c_{1:i-1}; A\right).
\]
For each occurrence of $X_t$, we compute the point estimate of STE as the log-likelihood difference under the “with-$Y$” and “without-$Y$” masks,
\begin{equation*}
S\widehat{T}E^{(t)}_{Y\to X}(\tau) \;=\; \sum_{i\in \mathcal{I}_X}  \Big[ 
    \log p_\theta\!\left(x_i \mid c_{1:i-1}; M\right) - \log p_\theta\!\left(x_i \mid c_{1:i-1}; \tilde M^{(\tau)}\right) 
\Big],
\end{equation*}
normalised per token to report bits per token via division by $|\mathcal{I}_X|\ln 2$. This equals the model-implied conditional information gain attributable uniquely to $Y$'s recent content, holding $X$'s own history and all other users $Z$ constant.

We estimate a dyadic, lag-specific STE by averaging across all target posts by $X$ that follow at least one post by $Y$ within the lag window $\tau$. Let $\mathcal{S}_X(\tau)$ denote the set of such target posts (indexed by $s$), and let $\widehat{T}^{(s)}_{Y\to X}(\tau)$ be the pointwise STE estimate for post $s$. The empirical dyadic STE is then
\[
S\widehat{T}E_{Y\to X}(\tau) \;=\; \frac{1}{|\mathcal{S}_X(\tau)|} \sum_{s \in \mathcal{S}_X(\tau)} \widehat{T}^{(s)}_{Y\to X}(\tau).
\]

\subsection*{Semantic Partial Information Decomposition}

\begin{figure}
    \centering
    \includegraphics[width=1\linewidth]{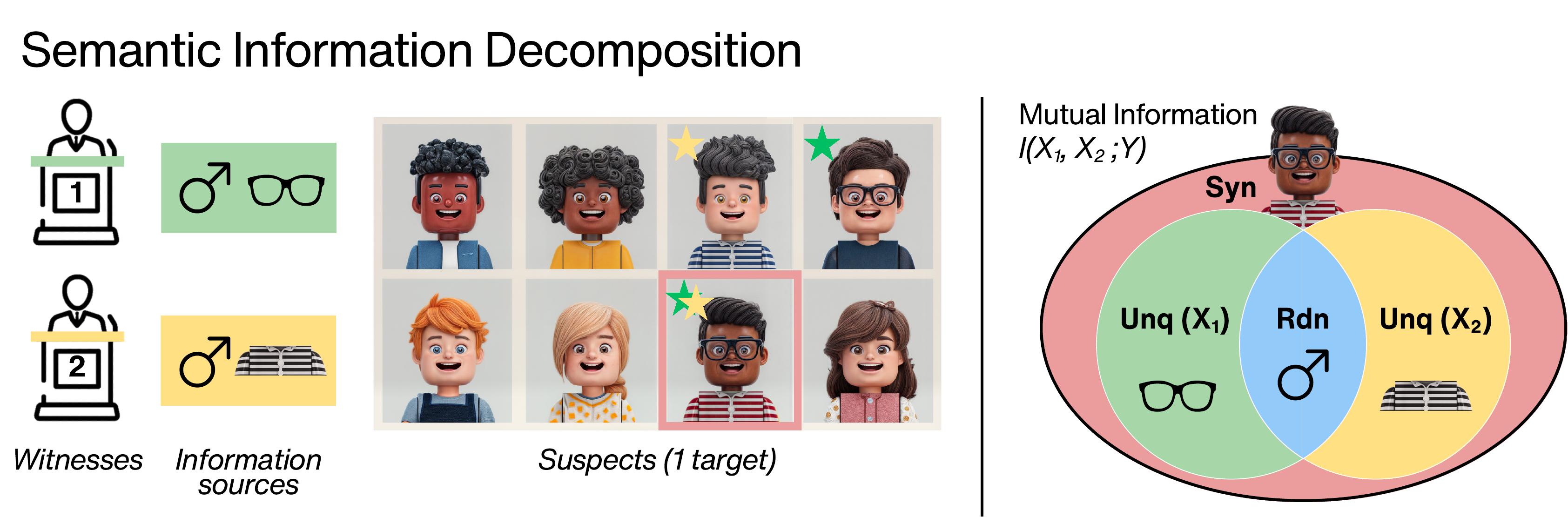}
    \caption{Semantic Partial Information Decomposition Illustration.}
    \label{fig:sid}
\end{figure}

As an illustration, consider a detective keen to reveal the identity of a thief from a line-up of suspects. Two witnesses (sources) come to provide information about the thief (target). The first witness indicates that the thief is a man and wears glasses, whilst the second also indicates that the thief is a man but wearing a striped shirt. Each source provided unique information (glasses; striped shirt), both provided redundant information (the thief is a man), but neither alone could conclusively identify the thief; only when both sources are considered synergistically does the identity emerge (Figure~\ref{fig:sid}). This example naturally scales to multiple sources of information beyond two.

To explicitly separate redundant, unique, and synergistic contributions, we apply a semantic partial information decomposition (SPID; \cite{liardi_null_2025, barrett_exploration_2015}) to the same model-induced mutual information terms as defined in STE. As in the STE estimator, we implement conditionalisation by selectively restricting the information available to the target sequence via attention masks. In the three-source setting considered here, this yields four masks that differ only in whether tokens of the target sequence $Y$ may attend to lag-window tokens from $X_1$ and/or $X_2$: (i) a baseline mask $\tilde M^{(\tau)}_{\emptyset}$ that blocks access to both sources, (ii) a mask $\tilde M^{(\tau)}_{X_1}$ that exposes $X_1$ while blocking $X_2$, (iii) a mask $\tilde M^{(\tau)}_{X_2}$ that exposes $X_2$ while blocking $X_1$, and (iv) the full mask $\tilde M^{(\tau)}_{X_1X_2}$, which exposes both sources. In all cases, the underlying token sequence, positions, and non-source context are held fixed.

Because the two source texts $X_1$ and $X_2$ are concatenated into a single context before the target, the autoregressive causal mask would ordinarily allow the later source to attend to the earlier one, introducing an asymmetry that has no counterpart in the SPID formalism (which treats sources symmetrically). We therefore impose an additional \emph{source-independence} constraint: in every mask, including $\tilde M^{(\tau)}_{X_1X_2}$, inter-source attention is blocked so that neither source's hidden representations are influenced by the other. This ensures that $I_\theta(Y;X_1)$ and $I_\theta(Y;X_2)$ are symmetric under permutation of sources, that $I_\theta(Y;X_1,X_2)$ reflects the joint contribution of two independently encoded sources, and that the resulting SPID atoms respect the symmetry axiom of \cite{williams_nonnegative_2010}.

All information quantities below are defined as log-likelihood gains relative to the same baseline mask $\tilde M^{(\tau)}_{\emptyset}$, which fixes the background information set against which redundancy, uniqueness, and synergy are assessed. For each occurrence of a target sequence $Y$, we therefore obtain the pointwise information gains
\[
\widehat{I}^{(t)}_\theta(Y;X_1)
=
\log p_\theta(Y\mid c_{1:n};\tilde M^{(\tau)}_{X_1})
-
\log p_\theta(Y\mid c_{1:n};\tilde M^{(\tau)}_{\emptyset}),
\]
\[
\widehat{I}^{(t)}_\theta(Y;X_2)
=
\log p_\theta(Y\mid c_{1:n};\tilde M^{(\tau)}_{X_2})
-
\log p_\theta(Y\mid c_{1:n};\tilde M^{(\tau)}_{\emptyset}),
\]
\[
\widehat{I}^{(t)}_\theta(Y;X_1,X_2)
=
\log p_\theta(Y\mid c_{1:n};\tilde M^{(\tau)}_{X_1X_2})
-
\log p_\theta(Y\mid c_{1:n};\tilde M^{(\tau)}_{\emptyset}).
\]
Averaging over all eligible target instances yields empirical estimates $\widehat{I}_\theta(\cdot)$.

\subsubsection*{Redundancy functions and PID atoms}
A PID is fully specified once a redundancy functional $\mathrm{Red}(X_1,X_2 \to Y)$ is chosen. Given any such redundancy definition, the remaining PID atoms follow algebraically:
\begin{align*}
\mathrm{Unq}(X_1 \to Y)
&=
\widehat{I}_\theta(Y;X_1)
-
\mathrm{Red}(X_1,X_2 \to Y), \\
\mathrm{Unq}(X_2 \to Y)
&=
\widehat{I}_\theta(Y;X_2)
-
\mathrm{Red}(X_1,X_2 \to Y), \\
\mathrm{Syn}(X_1,X_2 \to Y)
&=
\widehat{I}_\theta(Y;X_1,X_2)
-
\mathrm{Unq}(X_1 \to Y)
-
\mathrm{Unq}(X_2 \to Y)
-
\mathrm{Red}(X_1,X_2 \to Y).
\end{align*}
Different choices of redundancy functional therefore induce different PIDs over the same underlying information quantities.

\paragraph{Minimum Mutual Information (MMI).}
A simple and widely used redundancy functional is the Minimum Mutual Information (MMI) definition \cite{barrett_exploration_2015}, which identifies redundancy with the smaller of the two marginal information contributions:
\[
\mathrm{Red}_{\mathrm{MMI}}(X_1,X_2 \to Y)
=
\min\!\Big(
\widehat{I}_\theta(Y;X_1),
\widehat{I}_\theta(Y;X_2)
\Big).
\]
This choice is intuitive—redundancy cannot exceed what the weaker source provides—and yields a parsimonious decomposition, though it depends only on marginal information magnitudes and does not account for instance-level variation or sign structure in predictive effects.

\paragraph{Common Change in Surprisal (CCS).}
An alternative redundancy functional is given by the \emph{common change in surprisal} (CCS; \cite{ince_measuring_2017}), which defines redundancy in terms of local, instance-wise overlap in predictive effect. In our setting, the local overlap term for the $(t)$-th target is given by the model-induced co-information
\[
c_\theta^{(t)}(X_1;X_2;Y)
:=
\widehat{I}^{(t)}_\theta(Y;X_1)
+
\widehat{I}^{(t)}_\theta(Y;X_2)
-
\widehat{I}^{(t)}_\theta(Y;X_1,X_2).
\]
CCS redundancy is obtained by averaging this overlap only over instances for which the individual and joint surprisal changes agree in sign, thereby excluding ambiguous cases where sources act in opposing directions. Formally,
\[
\mathrm{Red}_{\mathrm{CCS}}(X_1,X_2 \to Y)
:=
\mathbb{E}_t\!\left[
c_\theta^{(t)}(X_1;X_2;Y)
\;\middle|\;
\operatorname{sign}\!\big(\widehat{I}^{(t)}_\theta(Y;X_1)\big)
=
\operatorname{sign}\!\big(\widehat{I}^{(t)}_\theta(Y;X_2)\big)
=
\operatorname{sign}\!\big(\widehat{I}^{(t)}_\theta(Y;X_1,X_2)\big)
\right].
\]
CCS thus provides a complementary redundancy functional that is sensitive to instance-level agreement in predictive effect, rather than relying solely on marginal information magnitudes.

\paragraph{Omission-based marginalisation.}
The formulation above describes marginalisation via attention masking, where sources remain in the token sequence but are hidden from the target's attention. An alternative, used in Experiment~3, is \emph{physical omission}: to compute $p(Y \mid X_1)$ without $X_2$, we remove $X_2$'s tokens entirely from the sequence, rebuild the input as $[X_1 \,\|\, Y]$, and evaluate with standard causal attention. Because the two sources in the SPID setting are independent texts with no intermediate responses, omission introduces no cascading changes to the remaining context, yielding the same conditional distributions as the masking formulation. The omission approach is computationally simpler (standard causal attention, no custom mask construction) and avoids any possibility of indirect information leakage. In all experiments reported here, the SPID results use the omission method.

\subsection*{Redundancy-Synergy Index}

For $n > 2$ sources, the number of partial information atoms grows with the Dedekind numbers (4, 18, 166 for $n = 2, 3, 4$), making a full PID computationally and conceptually unwieldy. The redundancy-synergy index (RSI; also called the whole-minus-sum measure; \cite{brenner_synergy_2000, schneidman_synergy_2003}) provides a scalar summary that captures the overall redundancy-synergy balance for any number of sources. RSI is defined as the difference between the joint mutual information and the sum of individual mutual informations,
\begin{equation*}
    RSI(X_1, \ldots, X_n; Y) = I(Y; X_1, \ldots, X_n) - \sum_{i=1}^{n} I(Y; X_i),
\end{equation*}
where positive values indicate synergy dominance (the joint contribution exceeds the sum of singletons) and negative values indicate redundancy dominance.

Each mutual information term is estimated as a log-likelihood difference following the same approach as SPID. For the joint term, we construct a sequence containing all $n$ sources concatenated with the target; for each singleton term $I(Y; X_i)$, we construct a sequence containing only source $X_i$ followed by the target (physically omitting all other sources). All log-likelihood differences are normalised by the target token count and converted to bits. Computing RSI for $n$ sources requires $n + 1$ forward passes (the joint plus each singleton, against a common baseline).

\subsection*{Experiment 1}

\paragraph{Conversation generation prompt.}
Conversations were generated via the OpenAI Batch API (\texttt{/v1/responses} endpoint) using GPT-5-nano with structured JSON output \cite{openai_gpt-5_2025}. Each request contained the following prompt template (with condition and topic substituted):

\begin{quote}
\small
Generate a 20-turn dialogue about \{topic\} between two agents who begin opposing each other’s positions. Each turn should have 2--3 sentences.

Condition: \{condition description\}

Cognitive rigidity is the difficulty in shifting from established patterns of thinking, feeling, or behaving when faced with new information or changing circumstances. Cognitive flexibility is the ability to adapt thinking and behavior in response to new information or environmental demands.

Return only valid JSON matching the schema.
\end{quote}

\noindent Condition descriptions were: ``Both speakers behave with cognitive rigidity. Speaker A starts.’’ (rigid-rigid); ``Both speakers behave with cognitive flexibility. Speaker A starts.’’ (flexible-flexible); ``Speaker A is cognitively rigid and Speaker B is cognitively flexible. Speaker \{A/B\} starts.’’ (rigid-flexible, alternating starting speaker); and ``Speaker A is cognitively flexible and Speaker B is cognitively rigid. Speaker \{A/B\} starts.’’ (flexible-rigid, alternating starting speaker). The two mixed conditions ensure that each cognitive-style direction (e.g., rigid$\rightarrow$flexible) is observed from both speaker positions (A$\rightarrow$B and B$\rightarrow$A), orthogonalising cognitive style and speaker position. The output was constrained to a JSON schema specifying \texttt{conversation\_id}, \texttt{condition}, \texttt{topic}, and an array of turns each containing \texttt{turn}, \texttt{speaker}, and \texttt{text} fields. Only API responses with HTTP status code 200 were retained.

\paragraph{Observation counts.}
The full dataset contained 44,997 turn-level STE observations across 3,000 conversations. After excluding turns 1--5 (lag window) and averaging to conversation-direction level, the analysis comprised 6,000 observations (1,000 each for r$\rightarrow$r and f$\rightarrow$f; 2,000 each for r$\rightarrow$f and f$\rightarrow$r).

\paragraph{Source type descriptive statistics.}
Table~\ref{tab:source_effect} reports mean STE by source cognitive style.

\begin{table}[h]
\centering
\caption{Semantic transfer entropy by source cognitive style.}
\label{tab:source_effect}
\begin{tabular}{lccccc}
\hline
Source Type & $M$ & $SD$ & $n$ & $F$ & $p$ \\
\hline
Rigid & 0.457 & 0.358 & 3000 & \multirow{2}{*}{41.07} & \multirow{2}{*}{$< .001$} \\
Flexible & 0.522 & 0.380 & 3000 & & \\
\hline
\end{tabular}
\end{table}

\paragraph{Effect sizes for all direction contrasts.}
Table~\ref{tab:si_direction_d} reports Cohen’s $d$ for all pairwise direction-type comparisons.

\begin{table}[h]
\centering
\caption{Cohen’s $d$ for all pairwise direction-type contrasts (Experiment~1).}
\label{tab:si_direction_d}
\begin{tabular}{lc}
\toprule
Contrast & Cohen’s $d$ \\
\midrule
f$\rightarrow$f vs f$\rightarrow$r & $-0.07$ \\
f$\rightarrow$f vs r$\rightarrow$f & $-0.24$ \\
f$\rightarrow$f vs r$\rightarrow$r & $-0.17$ \\
f$\rightarrow$r vs r$\rightarrow$f & $-0.18$ \\
f$\rightarrow$r vs r$\rightarrow$r & $-0.11$ \\
r$\rightarrow$f vs r$\rightarrow$r & $0.06$ \\
\bottomrule
\end{tabular}
\end{table}

\subsubsection*{Cross-Model Validation}

To assess whether the Experiment~1 STE findings depend on the choice of LLaMA~3.2-3B, we re-ran the full STE computation on all 3,000 conversations using Phi-3-mini-4k-instruct \cite{abdin_phi-3_2024} and Mistral-7B-v0.3 \cite{jiang_mistral_2023}. For each model, we report the two key quantities from the main analysis: mean STE for the f$\rightarrow$r and r$\rightarrow$f directions in mixed-dyad conversations, and the paired within-conversation contrast.

\begin{table}[h]
\centering
\caption{Cross-model replication of the directional STE asymmetry (Experiment~1). Mixed-dyad conversations only ($n = 2{,}000$ per direction per model). All values in bits per token.}
\label{tab:crossmodel_exp1}
\begin{tabular}{lcccc}
\hline
\textbf{Model} & \textbf{f$\rightarrow$r} & \textbf{r$\rightarrow$f} & \textbf{Paired $t$} & \textbf{$d$} \\
\hline
LLaMA~3.2-3B    & 0.513 & 0.450 & $-4.05^{***}$ & $-0.175$ \\
Phi-3-mini-4k   & 0.394 & 0.351 & $-3.56^{***}$ & $-0.147$ \\
Mistral-7B-v0.3 & 0.443 & 0.395 & $-3.50^{***}$ & $-0.149$ \\
\hline
\multicolumn{5}{l}{$^{***}p < .001$}
\end{tabular}
\end{table}

The directional asymmetry replicates across all three models and both architectures: f$\rightarrow$r consistently exceeds r$\rightarrow$f, with effect sizes in a narrow range ($d = -0.147$ to $-0.175$). The target type main effect observed with LLaMA (p = .032) did not replicate with Phi-3-mini (p = .836) or Mistral-7B (p = .544), indicating it reflects model-specific calibration rather than a robust signal. The source type effect is the replicable finding: flexible speakers generate higher directed predictive coupling regardless of model.

\subsection*{Experiment 2}

\begin{table*}[h]
    \centering
    \definecolor{lightgray}{gray}{0.9}
    \begin{tabularx}{\textwidth}{l X}
    \toprule
    \textbf{Strategy} & \textbf{Definition} \\
    \midrule
    Logical appeal & Persuading through logical reasoning, evidence, and statistics to show the donation’s tangible impact. \\
    Emotional appeal & Persuading by eliciting emotions such as empathy, guilt, anger, or personal involvement, often through stories. \\
    Credibility appeal & Establishing trust by citing credentials, organisational rankings, or external objective facts. \\
    Foot-in-the-door & Making a small initial request to increase compliance with a later, larger request. \\
    Self-modelling & Acting as a role model by expressing one’s own intention to donate or matching the persuadee’s donation. \\
    Personal story & Using personal narratives or lived experiences about donation or beneficiaries to motivate donation. \\
    Donation information & Providing specific procedural details about how the donation works, improving clarity and self-efficacy. \\
    Source-related inquiry & Asking whether the persuadee knows about the organisation involved in the task. \\
    Task-related inquiry & Asking for the persuadee’s opinions or interests about the organisation or donation task, without being personal. \\
    Personal-related inquiry & Asking about the persuadee’s personal experiences or background relevant to charity or children’s welfare. \\
    \bottomrule
    \end{tabularx}
    \caption{Persuasion strategies and definitions.}
    \label{tab:persuasion_strategies}
\end{table*}

\paragraph{Lag selection procedure.}
We selected the optimal lag window via a surrogate-corrected scan over candidate lags 1--25. For each lag $\ell$ and each conversation, we computed STE alongside 3 shuffled surrogates (random seed = 42), where surrogate conversations randomly permuted the temporal order of posts. For each lag, we recorded the median corrected TE (observed minus surrogate median) across conversations. We applied LOESS smoothing (span = 0.4) to the corrected TE trajectory and identified the elbow as the smallest lag reaching 95\% of the peak smoothed value. Beyond this point, incremental gains fell below 0.05 bits with declining signal-to-noise ratio (SNR $< 0.4$). This procedure selected a lag of 16 turns.

\paragraph{Conversation filtering.}
From the PersuasionForGood corpus, conversations were retained if they contained at least 2 posts, exactly 2 unique users (persuader = user 0, persuadee = user 1), and non-empty post content. All 300 conversations met these criteria.

\paragraph{Analysis sample.}
After filtering for finite STE, finite surprisal, and valid role assignment, the post-level dataset comprised 10,459 observations from 300 conversations.

\paragraph{Donation outcomes.}
Of 300 conversations involving 257 unique persuadees, 220 (73.3\%) resulted in a donation ($>$\$0). Median donation amount was \$0.25; mean donation was \$39.74.

\paragraph{Singular fit.}
The conversation-level random intercept in the main LME model converged to zero variance ($\sigma^2 = 0.000$), indicating that between-conversation variability was fully absorbed by the fixed effects. Results are therefore equivalent to a fixed-effects model with robust standard errors.

\paragraph{Linear mixed-effects model specification.}
The full model formula was:
\begin{quote}
\small
\texttt{te\_bits\_per\_token $\sim$ Condition + inv\_tokens + inv\_tokens\_sq + surprisal\_bits\_per\_token + appeal\_any + inquiry\_any + strategy\_logical\_appeal + strategy\_emotion\_appeal + strategy\_credibility\_appeal + strategy\_foot\_in\_the\_door + strategy\_self\_modeling + strategy\_personal\_story + strategy\_donation\_information + strategy\_source\_related\_inquiry + strategy\_task\_related\_inquiry + strategy\_personal\_related\_inquiry + (1~|~conversation\_id)}
\end{quote}

\noindent where \texttt{Condition} codes the target role (persuader vs persuadee), \texttt{inv\_tokens} and \texttt{inv\_tokens\_sq} are the inverse token count and its square (controlling for turn length), \texttt{surprisal\_bits\_per\_token} is the per-token surprisal under full context, and individual strategy indicators are binary flags for whether a given strategy was used in the conversation. The model was fitted with maximum likelihood estimation via \texttt{lme4}.

\paragraph{LME fixed effects.}
Table~\ref{tab:si_lme_coef} reports the full fixed-effects estimates.

\begin{table}[h]
\centering
\small
\caption{Fixed effects from the linear mixed-effects model (Experiment~2). $N = 10{,}459$ observations from 300 conversations. AIC = 43,085; BIC = 43,222.}
\label{tab:si_lme_coef}
\begin{tabular}{lrrr}
\toprule
Predictor & Estimate & SE & $t$ \\
\midrule
Intercept & 1.669 & 0.239 & 6.99 \\
Condition (persuadee) & $-2.238$ & 0.038 & $-59.09$ \\
Inverse tokens & 2.941 & 0.302 & 9.74 \\
Inverse tokens$^2$ & $-3.853$ & 0.308 & $-12.53$ \\
Surprisal (bits/token) & 0.106 & 0.012 & 8.63 \\
Appeal (any) & 0.232 & 0.235 & 0.99 \\
Inquiry (any) & 0.076 & 0.071 & 1.07 \\
Logical appeal & $-0.106$ & 0.040 & $-2.66$ \\
Emotional appeal & $-0.018$ & 0.039 & $-0.47$ \\
Credibility appeal & $-0.035$ & 0.066 & $-0.53$ \\
Foot-in-the-door & $-0.023$ & 0.040 & $-0.58$ \\
Self-modelling & $-0.061$ & 0.041 & $-1.48$ \\
Personal story & $-0.143$ & 0.047 & $-3.04$ \\
Donation information & $-0.072$ & 0.043 & $-1.69$ \\
Source-related inquiry & $-0.078$ & 0.050 & $-1.57$ \\
Task-related inquiry & $-0.001$ & 0.045 & $-0.03$ \\
Personal-related inquiry & 0.039 & 0.043 & 0.90 \\
\bottomrule
\end{tabular}
\end{table}

\paragraph{Donation logistic regression.}
The logistic model predicting donation ($>$\$0) was:
\begin{quote}
\small
\texttt{donated\_flag $\sim$ te\_standardized + donation\_information\_count + age\_z + agreeable\_z + care\_z + benevolence\_z + rational\_z}
\end{quote}
\noindent where \texttt{te\_standardized} is the Z-scored persuader$\rightarrow$persuadee STE, and psychological covariates from \cite{wang_persuasion_2020} were median-imputed and Z-scored. Table~\ref{tab:si_logistic} reports the full coefficients. A random-intercept variant (GLMER with \texttt{(1~|~persuadee\_id)}) yielded similar estimates (AIC = 347; donation-information: $\beta = 0.275$, $p = .018$).

\begin{table}[h]
\centering
\small
\caption{Logistic regression predicting donation ($>$\$0) from STE, strategy use, and psychological covariates (Experiment~2). $N = 300$. AIC = 345; BIC = 375.}
\label{tab:si_logistic}
\begin{tabular}{lrrrr}
\toprule
Predictor & $\beta$ & SE & $z$ & $p$ \\
\midrule
Intercept & 0.684 & 0.195 & 3.52 & $< .001$ \\
STE (standardised) & $-0.028$ & 0.137 & $-0.21$ & .837 \\
Donation-information count & 0.250 & 0.101 & 2.49 & .013 \\
Age & 0.157 & 0.147 & 1.07 & .284 \\
Agreeableness & 0.240 & 0.163 & 1.48 & .140 \\
Care & $-0.077$ & 0.145 & $-0.53$ & .593 \\
Benevolence & 0.263 & 0.144 & 1.83 & .067 \\
Rationality & 0.059 & 0.155 & 0.38 & .703 \\
\bottomrule
\end{tabular}
\end{table}

\paragraph{Strategy effects (full table).}
Table~\ref{tab:si_hedges} reports Hedges’ $g$ for all 10 persuasion strategies, comparing conversation-level persuader$\rightarrow$persuadee STE when a strategy was used versus not used. Welch’s $t$-tests and Mann--Whitney $U$ tests yielded consistent significance patterns.

\begin{table}[h]
\centering
\small
\caption{Strategy effects on persuader$\rightarrow$persuadee STE (Hedges’ $g$ with 95\% CI). Negative values indicate lower STE when the strategy is present.}
\label{tab:si_hedges}
\begin{tabular}{llrrr}
\toprule
Strategy & Class & $g$ & 95\% CI & $p$ \\
\midrule
Personal story & Appeal & $-0.51$ & [$-0.80$, $-0.22$] & $< .001$ \\
Credibility appeal & Appeal & $-0.45$ & [$-0.81$, $-0.10$] & .032 \\
Foot-in-the-door & Appeal & $-0.40$ & [$-0.64$, $-0.15$] & $< .001$ \\
Donation information & Appeal & $-0.39$ & [$-0.64$, $-0.14$] & .003 \\
Logical appeal & Appeal & $-0.30$ & [$-0.53$, $-0.08$] & .010 \\
Source-related inquiry & Inquiry & $-0.26$ & [$-0.49$, $-0.04$] & .023 \\
Emotional appeal & Appeal & $-0.14$ & [$-0.37$, $0.09$] & .224 \\
Self-modelling & Appeal & $-0.10$ & [$-0.34$, $0.14$] & .421 \\
Personal-related inquiry & Inquiry & $0.11$ & [$-0.13$, $0.35$] & .353 \\
Task-related inquiry & Inquiry & $0.09$ & [$-0.15$, $0.33$] & .446 \\
\bottomrule
\end{tabular}
\end{table}

\paragraph{Surprisal descriptive statistics.}
Median surprisal (bits per token under full context) was similar across roles: persuader $M = 3.94$ (SD = 0.65), persuadee $M = 3.99$ (SD = 0.80).

\subsubsection*{Cross-Model Validation}

To assess whether the Experiment~2 directional asymmetry depends on the choice of LLM, we re-ran the STE computation on all 300 PersuasionForGood conversations using Phi-3-mini-4k-instruct \cite{abdin_phi-3_2024} and Mistral-7B-v0.3 \cite{jiang_mistral_2023}. Supplementary Table~S4 reports mean STE in each direction and the paired within-conversation contrast.

\begin{table}[h]
\centering
\caption{Cross-model replication of the directional STE asymmetry (Experiment~2). All 300 conversations per model. All values in bits per token.}
\label{tab:crossmodel_exp2}
\begin{tabular}{lcccc}
\hline
\textbf{Model} & \textbf{Persuader$\rightarrow$Persuadee} & \textbf{Persuadee$\rightarrow$Persuader} & \textbf{Paired $t$} & \textbf{$d$} \\
\hline
LLaMA~3.2-3B    & 1.70 & 0.05 & $19.70^{***}$ & $1.14$ \\
Phi-3-mini-4k   & 1.70 & 0.11 & $39.36^{***}$ & $2.27$ \\
Mistral-7B-v0.3 & 2.61 & 0.12 & $48.98^{***}$ & $2.83$ \\
\hline
\multicolumn{5}{l}{$^{***}p < .001$}
\end{tabular}
\end{table}

The persuader$\rightarrow$persuadee asymmetry replicates across all three models with large effect sizes ($d = 1.14$–$2.83$), consistently exceeding the reverse direction by more than an order of magnitude. The null correlation between STE magnitude and persuasion success also replicated ($r = 0.05$–$0.11$, all $p > .06$), confirming that the structural asymmetry and its independence from outcome are not model-specific artefacts.

\subsection*{Experiment 3}

\subsubsection*{Cross-Model Validation}

To assess whether the Experiment~3 findings depend on the choice of LLM, we re-ran the STE computation on all 133 AnnoMI sessions using Phi-3-mini-4k-instruct \cite{abdin_phi-3_2024} and Mistral-7B-v0.3 \cite{jiang_mistral_2023}. Supplementary Table~S5 reports therapist-to-client STE by MI quality group and the corresponding effect sizes.

\begin{table}[h]
\centering
\caption{Cross-model replication of the therapeutic quality contrast (Experiment~3). All 133 AnnoMI sessions per model. All values in bits per token.}
\label{tab:crossmodel_exp4}
\begin{tabular}{lccccc}
\hline
\textbf{Model} & \textbf{High quality} & \textbf{Low quality} & \textbf{Welch $t$} & \textbf{$d$} & \textbf{Change talk $r$} \\
\hline
LLaMA~3.2-3B    & 0.76 & 1.62 & $2.40^{*}$  & $-0.80$ & $-.175^{*}$ \\
Mistral-7B-v0.3 & 0.87 & 1.75 & $2.43^{*}$  & $-0.71$ & $-.190^{*}$ \\
Phi-3-mini-4k   & 0.60 & 1.19 & $2.02$      & $-0.66$ & $-.190^{*}$ \\
\hline
\multicolumn{6}{l}{$^{*}p < .05$. High quality $n = 110$, Low quality $n = 23$.}
\end{tabular}
\end{table}

The direction of the effect is consistent across all three models: therapist-to-client STE is higher in low-quality than high-quality sessions in every case. The Phi-3-mini-4k result for the primary test is marginal ($p = .055$), likely reflecting the smaller absolute STE values produced by that model, but the directional asymmetry contrast reaches significance ($p = .035$, $d = -0.59$) and the change talk correlation replicates clearly ($r = -.190$, $p = .029$). The convergence across architectures and scales supports the conclusion that the quality contrast reflects a property of the sessions rather than an artefact of the estimation model.

\subsection*{Experiment 4}

\paragraph{Null model results (all atoms).}
Table~\ref{tab:si_null_atoms} reports the full comparison of real versus null (shuffled) SPID atoms. Unique information was positive for real pairings but negative in the null, indicating that shuffled premises were individually counter-predictive of the claim. Real pairings also showed lower synergy than null pairings (real: M = 0.32; null: M = 0.51; $t(1948) = -6.82$, $p < .001$, $d = 0.33$). This pattern follows from how the PID partitions $I(Y;X_1,X_2)$ under the MMI redundancy functional, which bounds redundancy by the smaller marginal source-target mutual information. In shuffled pairings, both marginals are near zero, so redundancy and both unique terms collapse and any residual joint predictive structure (generic regularities arising whenever two texts are concatenated) is absorbed into synergy by construction. In real pairings, premises carry substantial individual mutual information with the claim, most of which routes into redundancy (shared topical content) and unique contributions (premise-specific evidence), leaving synergy to capture only the super-additive contribution that emerges when premises combine semantically. The lower synergy in real pairings therefore does not imply weaker semantic synergy: the partition distributes genuine semantic coupling across all four atoms, while in shuffled controls the smaller, non-semantic joint signal is concentrated in synergy alone. Real synergy remained significantly positive, and the logic-gate unit test (Methods) confirms that SPID isolates genuine synergistic structure where it exists.

\begin{table}[h]
\centering
\small
\caption{Real vs null SPID atoms (bits per token). Negative unique values in the null indicate that random premises are counter-predictive of the claim.}
\label{tab:si_null_atoms}
\begin{tabular}{lrrrrr}
\toprule
Atom & Real $M$ & Null $M$ & $t$ & $p$ & $d$ \\
\midrule
Redundancy & 0.63 & 0.12 & 14.62 & $< .001$ & 1.42 \\
Unique $X_1$ & 0.20 & $-0.22$ & 15.25 & $< .001$ & 0.77 \\
Unique $X_2$ & 0.19 & $-0.23$ & 13.83 & $< .001$ & 0.75 \\
Synergy & 0.32 & 0.51 & $-6.82$ & $< .001$ & 0.33 \\
\bottomrule
\end{tabular}
\end{table}

\paragraph{CCS vs MMI comparison.}
CCS synergy ($M = 0.15$) was significantly lower than MMI synergy ($M = 0.32$), paired $t(324) = 11.45$, $p < .001$. MMI and CCS redundancy estimates were highly correlated ($r = .90$, $p < .001$).

\paragraph{Premise similarity.}
Semantic similarity between premise pairs (cosine similarity of all-MiniLM-L6-v2 embeddings) averaged $M = 0.42$ (SD = 0.16) across the 325 triplets.

\paragraph{Synergy and claim length.}
Synergy was negatively correlated with claim length ($r = -.21$, $p < .001$), suggesting that longer claims are easier to predict from individual premises alone.

\paragraph{Pointwise variability.}
Although synergy was positive on average, individual samples showed substantial variability, with some exhibiting negative pointwise synergy. This arises because we compute pointwise mutual information (per-token surprisal differences), which can be negative for individual observations even when the expected value is positive.

\subsubsection*{Masking versus Omission}

Attention masking and physical omission are appropriate to structurally different settings and should not be treated as interchangeable. In multi-turn STE, omitting a speaker’s tokens changes the positional encodings and hidden states of all downstream messages that attended to that speaker, so masking---which preserves the full sequence and severs only the relevant attention edges---is the correct choice. In the SPID setting, sources are independent and do not respond to one another, so omission is clean. Applying masking here is problematic because the full sequence $[X_1, X_2, Y]$ is substantially longer than the marginal sequences $[X_1, Y]$ or $[Y]$; masked-but-present source tokens shift $Y$’s positional encodings even when direct attention is blocked, inflating all log-likelihood estimates by an approximately constant offset.

We quantified this on all 325 AAE triplets by running both methods and comparing the resulting SPID atoms (Table~\ref{tab:leakage}). Individual mutual information terms were each inflated by 5.5--5.9 bits per token under masking. The inflation is approximately equal across $I(Y;X_1)$, $I(Y;X_2)$, and $I(Y;X_1,X_2)$, so it largely cancels in the synergy; but the residual distortion ($|\Delta\text{Syn}| = 0.45$ bits/token, $SD = 0.44$) reverses the sign in 59\% of samples (sign agreement: 41\%) and shifts mean synergy from $+0.32$ to $-0.09$. Had masking been used in Experiment~4, the central qualitative finding of positive synergy would have been reversed.

\begin{table}[h]
\centering
\caption{Mean information atom values (bits per token) under omission and masking on $N = 325$ AAE triplets.}
\label{tab:leakage}
\begin{tabular}{lccc}
\hline
Quantity & Omission & Masking & $|\Delta|$ \\
\hline
$I(Y;X_1)$       & 0.836 & 6.680 & 5.843 \\
$I(Y;X_2)$       & 0.819 & 6.506 & 5.688 \\
$I(Y;X_1,X_2)$   & 1.343 & 6.741 & 5.398 \\
Redundancy (MMI) & 0.633 & 6.357 & 5.724 \\
Synergy (MMI)    & $+$0.321 & $-$0.088 & 0.454 \\
\hline
\end{tabular}
\end{table}

This analysis does not carry over directly to STE, where both the ``with’’ and ``without’’ conditions operate on sequences of identical length and structure; the sequence-length artefact identified here cannot arise. The residual concern for STE is indirect leakage through intermediary speakers, which is structurally distinct from the positional artefact and likely smaller in magnitude.

\subsubsection*{Cross-Model Validation}

To assess whether the SPID findings from Experiment~4 depend on the specific LLM used, we replicated the full analysis (all 325 premise-claim triplets, plus null permutations) using Phi-3-mini-4k-instruct \cite{abdin_phi-3_2024} and Mistral-7B-v0.3 \cite{jiang_mistral_2023} alongside the original LLaMA~3.2-3B baseline. All models used the omission-based marginalisation method. Supplementary Table~S3 reports the key PID atoms for real and null (shuffled) premise pairings.

\begin{table}[h]
\centering
\caption{Cross-model comparison of SPID atoms on the AAE corpus ($n = 325$ triplets). Null values are from premise-claim pairs with premises drawn from different essays. All values in bits per token.}
\label{tab:crossmodel}
\begin{tabular}{lcccccc}
\hline
\textbf{Model} & \textbf{Params} & \textbf{Real syn} & \textbf{Real red} & \textbf{\% pos syn} & \textbf{Null red} & \textbf{Red ratio} \\
\hline
LLaMA~3.2-3B   & 3B   & 0.321 & 0.633 & 81.5\% & 0.121 & 5.2$\times$ \\
Phi-3-mini-4k  & 3.8B & 1.208 & 0.174 & 99.1\% & 0.005 & 38$\times$ \\
Mistral-7B-v0.3 & 7B  & 0.808 & 0.301 & 95.4\% & 0.015 & 20$\times$ \\
\hline
\end{tabular}
\end{table}

Across all three models and both architectures, mean synergy is positive and real redundancy substantially exceeds null redundancy. The real/null redundancy ratio ranges from 5$\times$ to 38$\times$, with the contrast growing at larger model scale---though this trend should not be over-interpreted as it may reflect differences in base entropy rates rather than sensitivity to topical relatedness. Absolute magnitudes are not comparable across models because MI values are denominated in each model’s own token probability space. The qualitative structure of the decomposition is consistent: premises from the same essay share more information with the claim than cross-essay premise pairs, regardless of the model used to estimate probabilities.

\newpage

\bibliographystyle{unsrt}
\bibliography{PSIDyn}

\end{document}